\documentclass[conference]{IEEEtran}
\IEEEoverridecommandlockouts
\usepackage{cite}
\usepackage{amsmath,amssymb,amsfonts}

\usepackage{graphicx}
\usepackage{textcomp}
\usepackage{xcolor}
\usepackage{subcaption}
\usepackage{caption}
\usepackage{algorithm}
\usepackage{algpseudocode}
\usepackage{mathtools}
\usepackage{hyperref}

\def\BibTeX{{\rm B\kern-.05em{\sc i\kern-.025em b}\kern-.08em
    T\kern-.1667em\lower.7ex\hbox{E}\kern-.125emX}}

\begin{document}

\title{Autonomous Vision-based UAV Landing with Collision Avoidance using Deep Learning
\\
{\footnotesize Tianpei Liao, Amal Haridevan, Yibo Liu, Jinjun Shan}

}

\makeatletter
\newcommand{\linebreakand}{%
  \end{@IEEEauthorhalign}
  \hfill\mbox{}\par
  \mbox{}\hfill\begin{@IEEEauthorhalign}
}
\makeatother

\author{\IEEEauthorblockN{1\textsuperscript{st} Tianpei Liao}
\IEEEauthorblockA{\textit{Department of Electrical Engineering and Computer Science} \\
\textit{York University}\\
North York, Canada \\
liaot3@my.yorku.ca}
\and
\IEEEauthorblockN{2\textsuperscript{nd} Amal Haridevan}
\IEEEauthorblockA{\textit{Department of Earth and Space Science and Engineering} \\
\textit{York University}\\
North York, Canada \\
amaldev@my.yorku.ca}
 \linebreakand
\IEEEauthorblockN{3\textsuperscript{rd} Yibo Liu}
\IEEEauthorblockA{\textit{Department of Earth and Space Science and Engineering} \\
\textit{York University}\\
North York, Canada \\
yorklyb@yorku.ca}
\and
\IEEEauthorblockN{4\textsuperscript{rd} Jinjun Shan}
\IEEEauthorblockA{\textit{Department of Earth and Space Science and Engineering} \\
\textit{York University}\\
North York, Canada \\
jjshan@yorku.ca}
}

\maketitle

\begin{abstract}
The autonomous vision-based Unmanned Aerial Vehicles (UAVs) landing is an adaptive way to land in special environments such as the global positioning system denied. There is a risk of collision when multiple UAVs land simultaneously without communication on the same platform. This work accomplishes vision-based autonomous landing and uses a deep-learning-based method to realize collision avoidance during the landing process. Specifically, the landing UAVs are categorized into level I and II. The deep learning method will be implemented for level II UAV. The YoloV4 deep learning method will be implemented by the Level II UAV to achieve object detection of Level I UAV. Once the Level I UAV's landing has been detected by the onboard camera of Level II UAV, it will move and land on a relative landing zone beside the Level I UAV. The experiment results show the validity and practicality of our theory.

\end{abstract}

\section{Introduction}
Unmanned Aerial Vehicle (UAV) development has shown valuable potential in the delivery market. The companies like Amazon and Alibaba develop a considerable interest in drones delivery and have started competing about testing drones to deliver packages \cite{r1}. Various UAVs have different purposes including transportation of food, medical supplies, and packages. A key aspect of drones delivery is autonomous landing. Recently, researchers have shown an increased interest in autonomous UAV landing using fiducial markers. The purpose of using fiducial markers is to estimate the pose of the vehicle by obtaining the six-degrees of freedom \cite{r2}. A number of techniques, such as Apriltag \cite{r2} and ARTtag \cite{r5}, have been developed to adapt and explore autonomous landing.

In the situation that two different UAVs can not communicate and share information, that is having no vehicle safety communication (VSC) \cite{r4}, there is an urgent need to address the safety problems caused by the collision during autonomous landing. 
 \begin{figure}[htbp]
\centerline{\includegraphics[scale=.3]{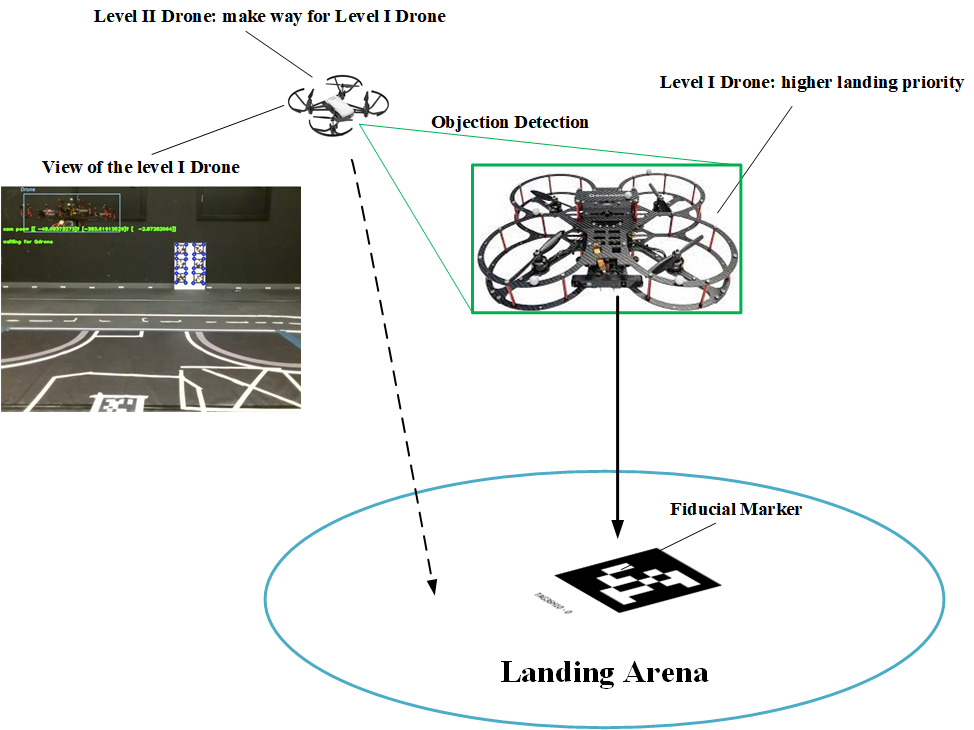}}
\caption{Diagram of proposed framework}
\label{fig}
\end{figure} Due to the complexity of the business model UAVs, the conflict of “Free Flight” \cite{r3} is possible to occur and yields the dangerous movement. For the large-scale UAV, one of the greatest challenges is to estimate the spatial relation with another close UAV so that it can archive a safe landing path. Depending on the functionality of UAVs, the different priority of vehicles needs to be assigned if they are landing closely and recently.

In this research, we purpose a strategy to resolve the risk of collision when two different levels of UAV landing on close paths. The Apriltags are used for navigation and estimating position due to its efficiency, less false positive rate, and robustness \cite{r2}. The two different levels of UAV are categorized into Level I and II UAV and they are landing on close platforms. To clarify the problem of having no VSC \cite{r4}, the vision-based collision avoidance method is introduced. An inexpensive detection algorithm is implemented to archive real-time decision-making. Moreover, the YoloV4 \cite{r7} deep learning approach is adopted on Level II UAV to obtain further in-depth information on object detection. The Non-Maximum Suppression (NMS) \cite{r8,r11} is utilized to avoid multiple bounding boxes so that it raises detection accuracy. The Level I UAV is labeled by the bounding box on the view of Level II UAV. The path of the bounding box can be understood by Level II UAV to determine if Level I UAV has finished landing. The Level II UAV will safely move to its landing zone after it determines the landing of Level I UAV.

\section{Proposed Method}
\subsection{Estimation of position}
The recent Apritag detection has been improved on detection speed and localization accuracy by implementing the continuous boundary segmentation algorithm \cite{r2}. To estimate its 3D coordinate in the world coordinate system, the tag position is required to be obtained first. In the situation that the Level II UAV always sees the Apriltags, the 3D reference points of each Apriltag 4-corners and their 2D projection need to be resolved, which is considered as “Perspective-n-Point problem” (PnP). The solving PnP method is based on the calibrated camera, so it demands for the camera intrinsic matrix $A$ after the camera calibration. The closed-form solution is used to gain the intrinsic parameters of $A$ in camera calibration \cite{r6}. A primary concern of solving the PnP problem is coordinate transformation. The system consists of the pixel, image, camera, and world coordinate system. To solve the PnP problem and regulate different coordinate systems, the relative method \cite{r9} is used   \begin{equation} s\: p = A \: [R \: |\: t ]\: P_w  \end{equation}
where $A$ is a $3\times3$ camera intrinsic matrix. The join matrix $[R | t]$ consists of rotation $R$ and translation $t$, which is obtained from Apriltag's coordinates and orientation. $P_w$ represents the 3D point with respect to world coordinate system. The corresponding 2D pixel with respect to image coordinate system denotes as $p$ and $s$ is the scaling factor \cite{r9}.

\subsection{YoloV4 dataset cloud training}
\begin{figure}[htbp]
\centerline{\includegraphics[scale=.25]{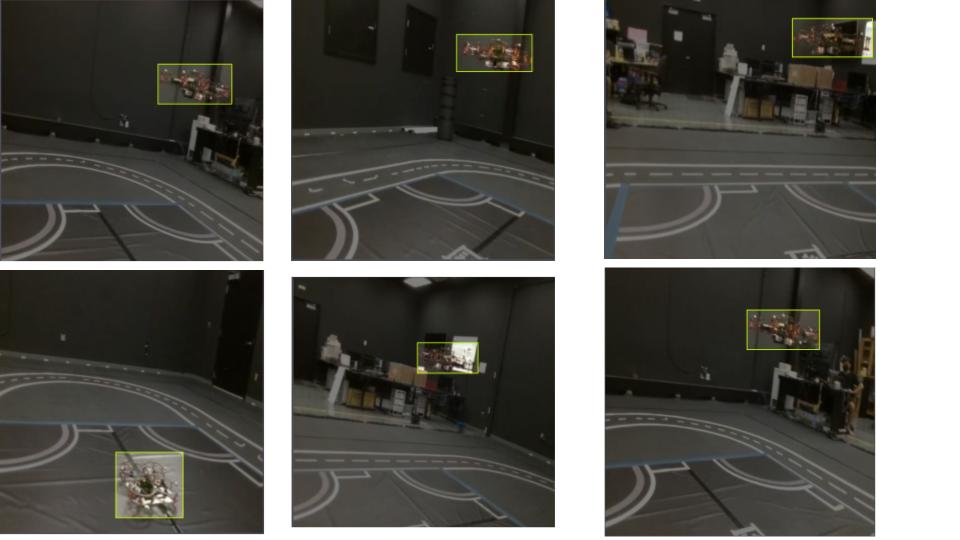}}
\caption{Example of labeling}
\label{fig}
\end{figure}
One of the concern about real-time detectors is the tradeoff between Graphics Processing Units(GPU) usage and detection accuracy. The improved pipeline computation implemented in YoloV4 produces high-quality object detection in real-time \cite{r7}. Therefore, YoloV4 is applied for our collision avoidance strategy to compensate for the massive memory usage during the real flight time. The advantage of YoloV4 provides faster FPS and a more accurate Average Precision detector \cite{r7}.

Figure 2 shows the example of labeling Level I UAV in the training dataset. The output images will be cropped into the customized size that matches your detector algorithm. Cloud dataset training is an innovative and convenient training method. It does not require sophisticated configuration with hardware since the environment has been built up in the cloud server. The Google Colab Pro \cite{r10} was recently introduced by Google that connects with the super engine at the backend Google cloud server and extremely increases the training process.

\subsection{Collision avoidance}

The Level I UAV has no awareness of Level II UAV. It will perform the automatic landing on the Apriltag that is placed on the ground. To gain the path of Level I UAV’s landing path on Level II’s onboard camera, we calculate the gap of the previous and current bounding box with respect to image coordinate system. The procedure of collision avoidance requires the detection of Level I complete landing.
\begin{figure}[htbp]
\centerline{\includegraphics[scale=.38]{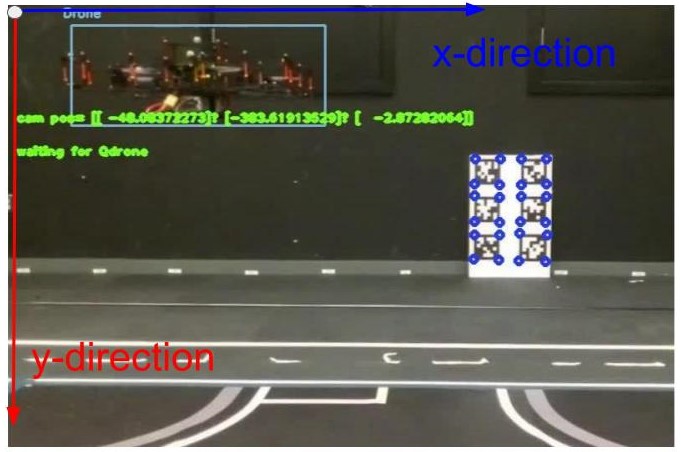}}
\caption{Illustration of image coordinate system}
\label{fig}
\end{figure}

Figure 3 shows the 2D image coordinate system where the top left corner is the origin. The iterative image will be processed into a 4-dimensional blob after the detection. The blob is denoted as $[x , y, w, h]$. The $x, y$ are the coordinates of the center of bounding boxes. The $w, h$ are the width and height of bounding boxes. One blog is considered as the collection of images with the same width, height, and depth. 
To increase detection accuracy during real flight time, we use Non-Maximum Suppression to filter out some of the bounding boxes that have poor accuracy. The filter consists of two parts. In order to minimize the candidates of filtering. First, simply select the confidence score that is higher than 0.5 in the set of predictions, which yields a new set. Let $P$ represents the set of all possible prediction, and $P’$ be the new filtered set. 
$P' = \{p \in P \mid c\_score(p) > 0.5\} $

\begin{figure}[htbp]
\centerline{\includegraphics[scale=.22]{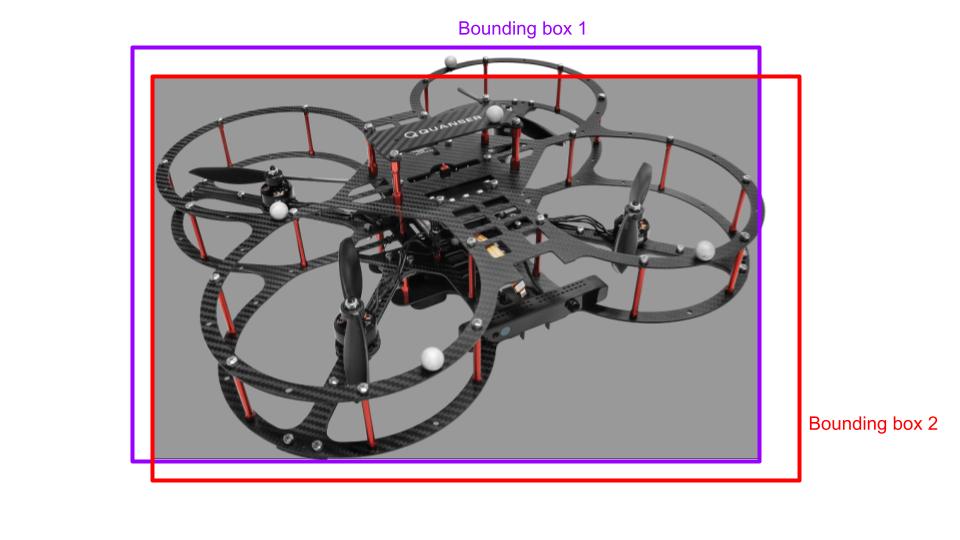}}
\caption{Intersection area between bounding box 1 and 2}
\label{fig}
\end{figure}

The highest confidence score from $P'$ will be selected and then calculate the Intersection over Union (IoU) value with all other elements from $P’$. The IoU value is the intersection area between the highest confidence score and one of the selected bounding boxes in $P’$ divided by the union areas. The Figure 4 shows the grey area as the intersection area of bounding boxes. If the IoU exceeds the IoU threshold that we define, then the selected bounding box will be removed from $P’$. Repeat the process until there is no prediction left in $P’$. The IoU threshold is $0.4$ in our experiment.

\begin{figure}[ht]
  \subfloat[Bounding boxes without NMS]{
	\begin{minipage}[c][0.9\width]{
	   0.23\textwidth}
	   \centering
	   \includegraphics[width=1\textwidth]{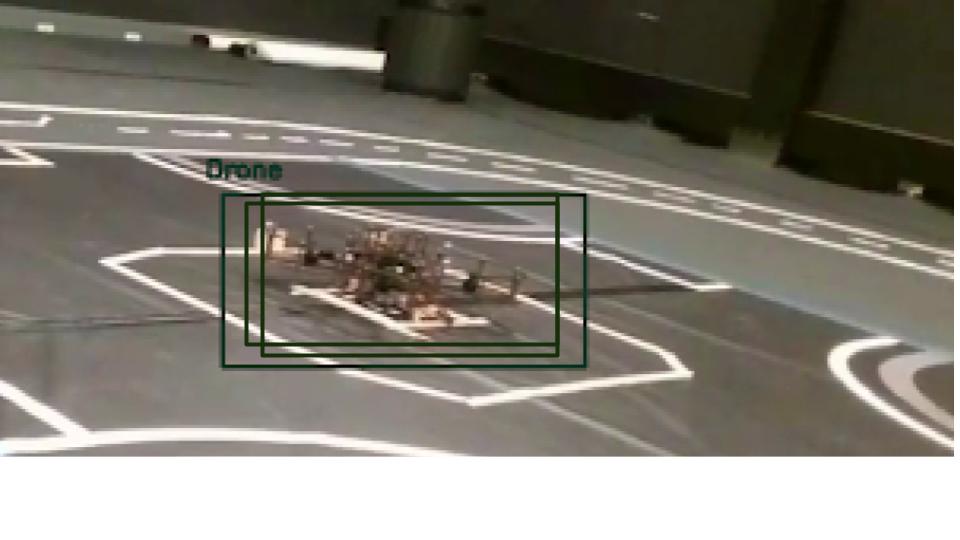}
	\end{minipage}}  \subfloat[Bouding box with NMS]{
	\begin{minipage}[c][0.9\width]{
	   0.23\textwidth}
	   \centering
\includegraphics[width=1\textwidth]{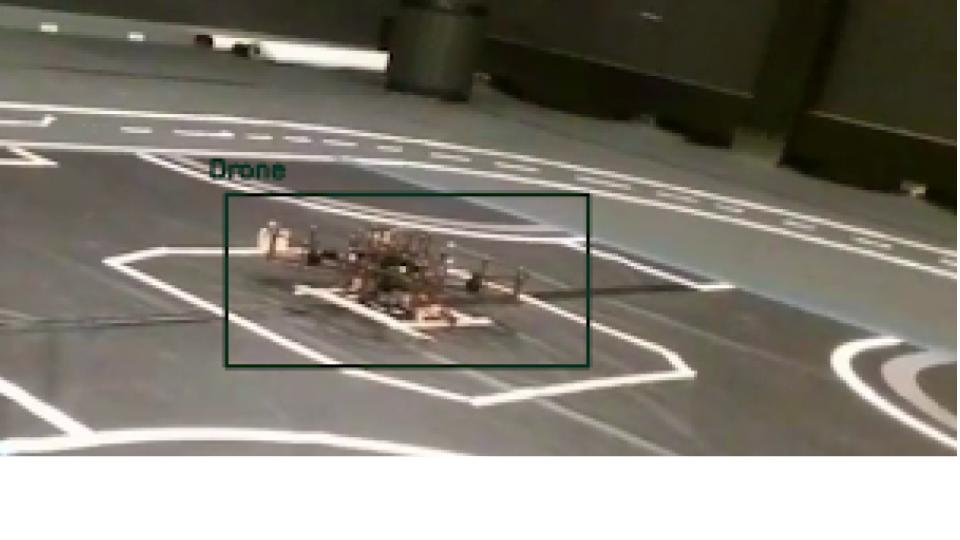}
	\end{minipage}}
\caption{Difference between absence or presence of using NMS}
\end{figure}
Figure 5(a) shows the multiple bounding boxes with various confidence scores before NMS filtering. Figure 5(b) shows the result after the NMS filtering. 

The bounding box coordinate in real time is represented by ${(x_t, y_t) } $. The function $f(t)$ is the distance difference of current and previous bounding box
\begin{equation}
f(t) = \sqrt{(x_{t+\Delta t} - x_t)^2 + (y_{t+\Delta t} - y_t)^2} \end{equation} where $(x_t,y_t)$ is the current bounding box coordinate and $(x_{t+\Delta t},y_{t+\Delta t})$ is the previous bounding box coordinate. The change with respect to image coordinate in x-axis might be slightly since the landing mostly affects the bounding box in y-axis. The sampling time of vehicle is represented as $\Delta t$. The landing threshold is denoted as $\sigma$ which demonstrates the minimum transition of Level I UAV landing on Level II UAV image plane. If $f(t) > \sigma$, the Level II UAV determines that Level I UAV has completed landing.

\begin{figure}[htbp]
\centerline{\includegraphics[scale=.4]{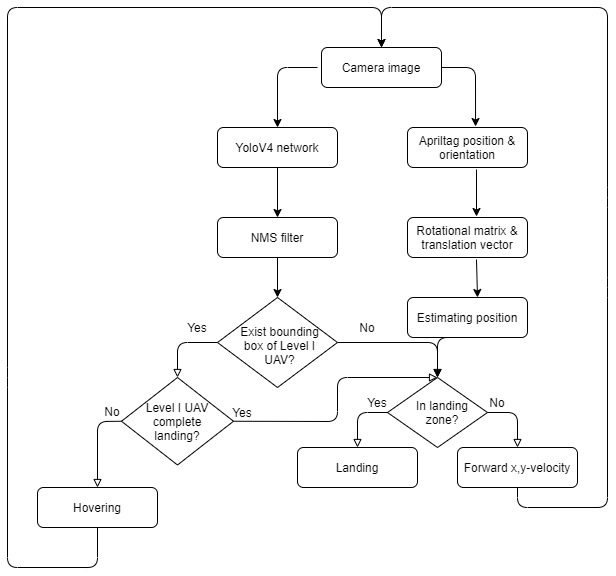}}
\caption{Proposed procedure of autonomous landing collision avoidance}
\label{fig}
\end{figure}
Figure 6 shows the procedure of Level II UAV autonomous landing collision avoidance. If Level II UAV detects the existence of Level I UAV, it will hover until the Level I UAV concludes landing. Otherwise, it keeps tracking the iterative images and moves to the landing zone.
\section{Experiment}

Level I UAV is presented by Qdrone from Quanser which has Omnivision OV7251 as the down camera. It is considered a heavy-duty UAV that has higher priority due to its payload and heavyweights. Level II UAV is presented by DJI Tello which has the front camera with 720HD transmission. It is considered an agile UAV with lower priority and needs to wait for Level I UAV’s landing. Since both Level I and II UAV’s landing zone are close, the autonomous system is controlled by the pre-trained dataset and machine learning algorithms. The pose estimation and automatic landing are based on fiducial markers.

In the experiment, there are 6 Apriltag used for level II UAV pose estimation, whose family is 36h11 and 0-5 id. There is a total of 24 n points from Apriltags. The 3D points of Apriltag with respect to world coordinate origin needs to be assigned first. For Level I UAV autonomous landing, it will capture the Apriltag (id:6) by its down camera within the limited range, and land on the Apriltag (id:6).

\begin{figure}[htbp]
  \subfloat[State 1: Waiting]{
	\begin{minipage}[c][1\width]{
	   0.25\textwidth}
	   \centering
	   \includegraphics[width=1\textwidth]{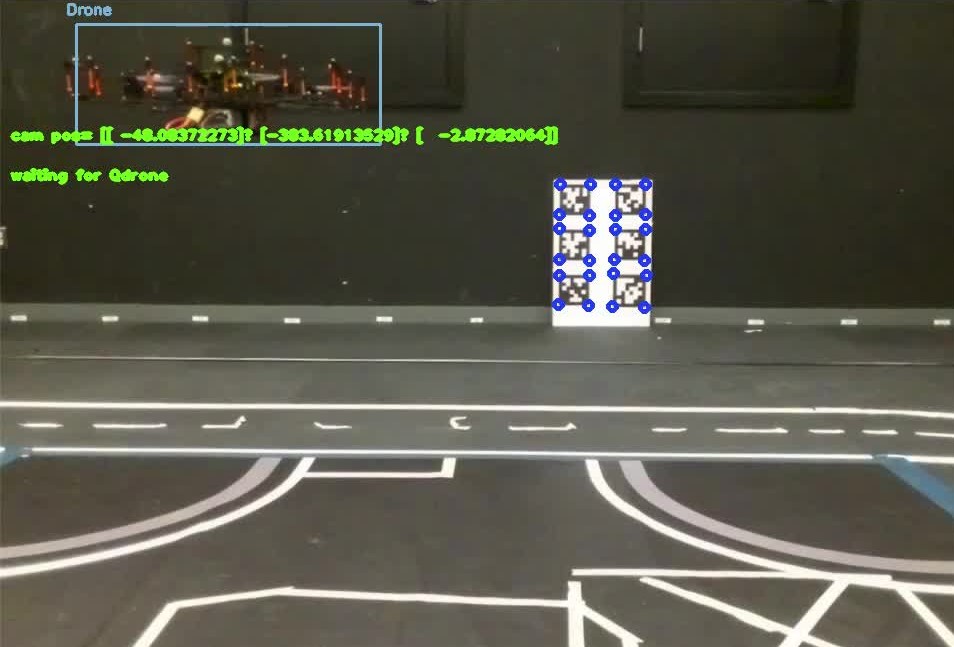}
	\end{minipage}}  \subfloat[State 2: Moving]{
	\begin{minipage}[c][1\width]{
	   0.25\textwidth}
	   \centering
\includegraphics[width=0.9\textwidth]{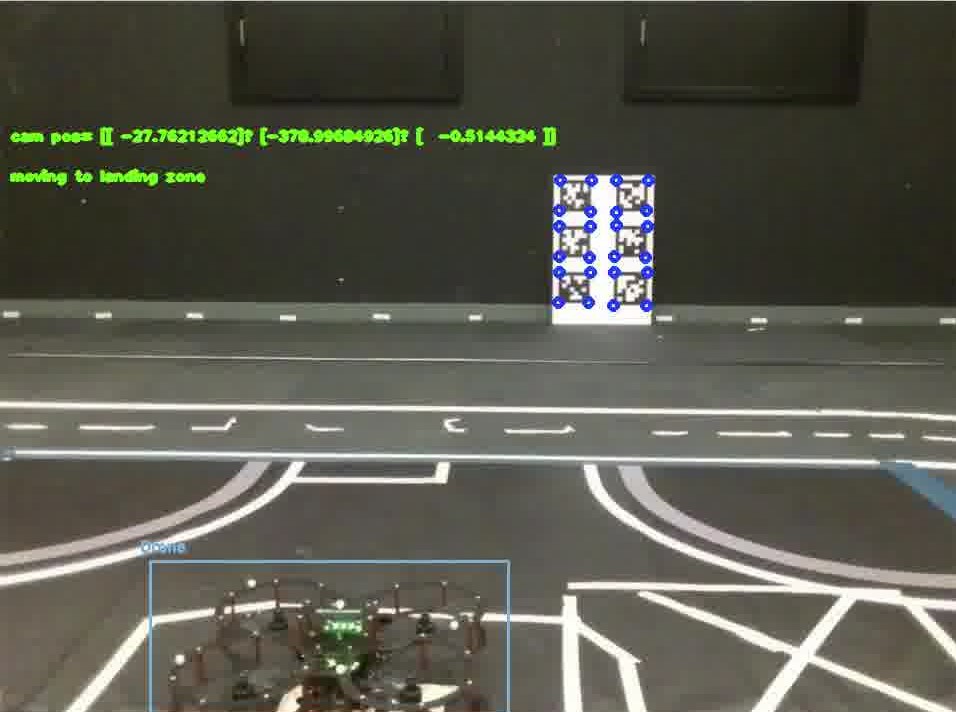}
	\end{minipage}}
\caption{2 States of Level II UAV}
\end{figure}

Figure 7 shows 2 different states of Level II UAV. It switches from normal flight to State 1 if it is seeing Level I UAV. This will forward 0 cm/s velocities to the remote control of the motors, which means hovering and waiting. If $f(t)$ in Equation (2) exceeds landing threshold $\sigma$, it will consider that the Level I UAV accomplishes landing and switch to State 2. The landing threshold of the experiment $\sigma$ is 150 in pixels. State 2 is showing in Figure 7(b).

\begin{figure}[htbp]
\centering
\begin{subfigure}[b]{0.3\textwidth}
   \includegraphics[width=1\linewidth]{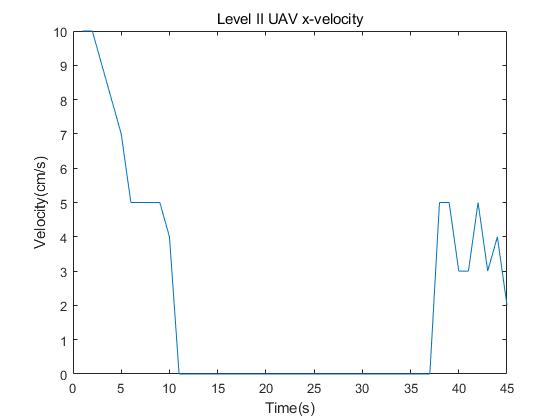}
   \label{fig:Ng1} 
\end{subfigure}

\begin{subfigure}[b]{0.3\textwidth}
\includegraphics[width=1\linewidth]{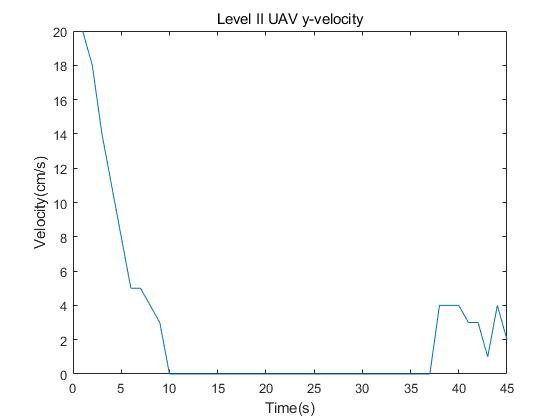}
\label{fig:Ng2}
\end{subfigure}
\caption{Level II UAV x,y-velocity}
\end{figure}

Figure 8 shows the x,y-velocity with respect to world coordinate in cm/s. The positive y-direction is toward the Apriltags and the positive x-direction is to the right when the Level II UAV front camera is facing the Apriltags. The value of velocity is proportional to the difference between current and expected position. So it will become slower when it is approaching the expected landing zone as the difference is decreasing. When the Level II UAV is taking off, it is moving to expected landing zone by forwarding x,y-velocity to remote control. After 11 seconds, the bounding box of Level I UAV has been generated by object detection. It forwards 0 cm/s velocity to all the motors as it is waiting for Level I UAV. After 37th second, it detects the Level I UAV complete landing, so it moves to the landing zone and land. To compensate  for  the  latency  of  implementing deep  learning  object  detection, there is a time delay of one second after sending the velocity to remote control. The video of our experiment is shown here \url{https://youtu.be/AtY4MuwV8tM}.

\section{Conclusion}
This project was undertaken to evaluate the risk of collision under multiple UAVs which are having no vehicle safety communication (VSC) \cite{r4} and design a vision-based collision avoidance method during the autonomous landing of 2 different levels of UAV. The present study lays the groundwork for future research into the real-time decision-making of collision avoidance using the object detection bounding boxes. Further studies need to be carried out in order to validate the prediction of object movement. It will adapt more complex trajectories and compensate the bounding boxes that contain low confidence scores.

\bibliographystyle{IEEEtran}
\bibliography{Reference.bib}
\end{document}